\documentclass{article}

\PassOptionsToPackage{numbers,sort&compress}{natbib}
\usepackage[preprint]{neurips_2026}
\workshoptitle{Who Verifies the Agents? Toward Reliable Agent Development}
\makeatletter
\renewcommand{\@noticestring}{Workshop on Who Verifies the Agents? Toward Reliable Agent Development, NeurIPS 2026.}
\makeatother

\usepackage[utf8]{inputenc}
\usepackage[T1]{fontenc}
\usepackage{hyperref}
\usepackage{url}
\usepackage{xurl}
\usepackage{xspace}
\usepackage{booktabs}
\usepackage{amsmath}
\usepackage{amssymb}
\usepackage{graphicx}
\usepackage{microtype}
\usepackage{xcolor}
\usepackage{tabularx}
\usepackage{array}
\usepackage{multirow}
\usepackage{nicefrac}
\usepackage{placeins}
\usepackage{needspace}

\definecolor{paperblue}{HTML}{276FBF}
\definecolor{paperteal}{HTML}{1B998B}
\definecolor{paperorange}{HTML}{E07A3F}
\definecolor{paperred}{HTML}{C44536}
\hypersetup{colorlinks=true,citecolor=paperblue,linkcolor=paperblue,urlcolor=paperblue}

\newcommand{\system}{\textsc{RefinementGuard}\xspace}
\newcommand{\localize}{\texttt{localizeBug}\xspace}
\newcommand{\validatepatch}{\texttt{validatePatch}\xspace}
\newcommand{\bzero}{\textsc{B0}\xspace}
\newcommand{\bone}{\textsc{B1}\xspace}
\newcommand{\btwo}{\textsc{B2}\xspace}
\newcommand{\bthree}{\textsc{B3}\xspace}
\newcommand{\qcert}{qualification certificate\xspace}
\newcommand{\yes}{\textcolor{paperteal}{\ensuremath{\checkmark}}}
\newcommand{\no}{\textcolor{paperred}{--}}
\newcolumntype{Y}{>{\raggedright\arraybackslash}X}


\title{RefinementGuard:\\ Counterexample-Directed Semantic Qualification for Hidden Agent-Tool Execution}

\author{
  Adhyayan Veer Singh\\
  {\normalfont The Quarry Lane School}
  \And
  Fanjiang Ye\\
  {\normalfont Rice University}
  \And
  Yuke Wang\\
  {\normalfont Rice University}
}

\begin{document}
\maketitle

\begin{abstract}
Tool runtimes can lower an agent-selected operation into a hidden parallel graph, reducing latency and context while risking semantic mistranslation. We introduce \system: translation validation for hidden, effectful agent--tool lowerings, coupled to artifact-bound, fail-closed runtime admission. Qualification compares a candidate graph with an independent reference on isolated state using explicit parent-operation obligations. We evaluate two repository-level contracts---bug localization and patch validation---on controlled repositories. Among 100 activated, non-equivalent development/qualification faults adjudicated by one human, parent-semantic validation rejects 49/100 (operator-cluster bootstrap 95\% CI [33.7\%, 64.8\%]). The objective is selective admission, not maximum mutation rejection: exact equality rejects 100/100 faults but also all 60 valid variants, whereas obligation-based validation accepts 60/60; its 51 residual misses identify concrete specification gaps. On a frozen disjoint input split, the held-out \btwo--\bthree result is localization-only because 21 patch mutants fail structurally; \bthree rejects 38/56 localization defects. In a controlled asynchronous-delay regime, the localization serial/guarded median-latency ratio is 4.50$\times$, and hidden execution exposes 74.3\% less parent-visible context than visible execution. Patch validation shows no demonstrated speedup; zero-delay runs expose overhead. End-to-end qualification thus detects lower-layer misses while preserving localization's parallelism.
\end{abstract}

\section{Introduction}

High-level tool interfaces let an agent request ``localize the bug'' or ``validate this patch'' while the runtime privately fans work into searches, reads, tests, and analyses. Independent branches run concurrently, and intermediate outputs need not consume the agent's context. The interface stays stable as the implementation becomes faster and more context-efficient \citep{kim2024llmcompiler}.

This abstraction creates a correctness risk. A localization graph can omit an applicable error-search branch yet remain acyclic, type-correct, within budget, and locally valid at every node that executes. A patch graph can run well-formed checks against the wrong repository revision or suppress a mandatory failure. The agent selected the right operation, but the harness implemented the wrong one. These are agent--tool semantic mistranslations, not malformed calls or ordinary model errors.

\textbf{Key contribution.} \system brings translation validation to hidden, effectful agent--tool lowerings and connects the result to artifact-bound, fail-closed runtime admission. Qualification executes a developer-authored candidate and an independently invoked reference on isolated repository states, then compares their normalized parent observations using explicit obligations. Parent schemas (\bzero), graph structure (\bone), and node-local contracts (\btwo) retain their narrower roles; parent-semantic validation (\bthree) checks status, failure, effects, visibility, budgets, grounding, provenance, branch completeness, ordering, mandatory checks, and repository revision. Mutation labels are unavailable to \bthree. A qualification certificate binds the resulting evidence to the exact contract, graph, reducer, tools, runtime, obligations, environment, and input domain.

The admission step is essential: evidence for one concrete graph says little after its reducer, tools, obligations, environment, or domain changes. Before scheduling, the guard recomputes every binding. A missing, stale, mismatched, quarantined, or out-of-domain certificate keeps \texttt{candidateStarted=false} and invokes the reference implementation when available. Qualification-time semantic codes and runtime admission codes remain separate.

\begin{figure}[t]
  \centering
  \includegraphics[width=\linewidth]{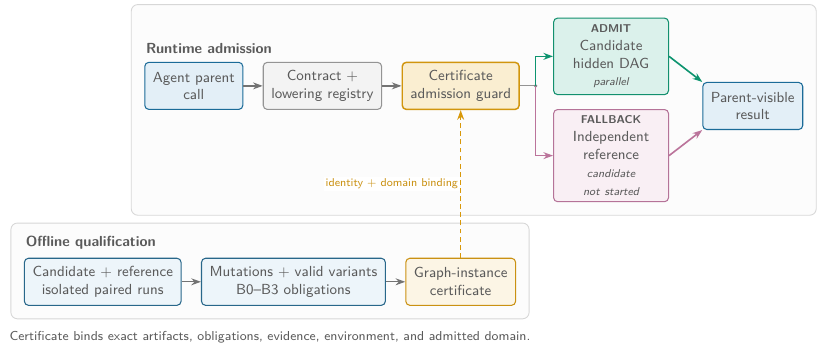}
  \caption{\system separates offline translation validation from online admission. Qualification compares a hidden lowering with an independent reference on isolated state. The resulting certificate binds the validated artifacts, obligations, environment, and input domain; any runtime mismatch selects reference fallback before the candidate starts.}
  \label{fig:architecture}
\end{figure}

Figure~\ref{fig:architecture} shows how offline qualification and online admission meet at the certificate boundary. Paired candidate--reference runs establish evidence for an exact lowering and domain; the runtime guard then either admits that hidden DAG or selects the independent reference before candidate execution begins.

The evaluation centers on three questions: \emph{Do lower layers miss parent-semantic defects? Does \bthree balance defect sensitivity with tolerance for valid implementation variation? Does hidden execution retain latency and context benefits?} The answers are bounded but encouraging. Across 100 non-equivalent development/qualification faults in two repository contracts, \bthree rejects 49, including 37/88 accepted by every lower layer. This is not a maximum-rejection objective: exact equality rejects all 100 faults but also all 60 valid variants, whereas \bthree accepts 60/60 and leaves 51 counterexamples as an explicit obligation-refinement queue. On a disjoint frozen input split, held-out semantic evidence is localization-only and yields 38/56 \bthree rejections. In the controlled asynchronous-delay regime, localization has a 4.50$\times$ case-level median serial/guarded latency ratio and 74.3\% less parent-visible context than visible execution; patch validation has no demonstrated latency gain. These results establish finite qualification for the two disclosed contracts and domains, not universal refinement or downstream live-agent gains.

The paper's contribution is therefore a systems boundary: validate the translation from a selected semantic tool to its hidden effectful implementation, then admit only the exact artifact and domain supported by that evidence. The evaluation supplies counterexamples, valid-variation controls, clustered inference, and runtime evidence for that boundary.

\section{Problem and system model}
\label{sec:model}

\paragraph{Semantic call and lowering.}
Let $C$ be a registered parent contract with input $x$, and let $K_C$ be an independent canonical implementation. The candidate lowering is a developer-authored graphlet $G_C(x)$ whose nodes invoke tools and whose edges encode data dependencies. An asynchronous executor $E$ schedules value-ready nodes under resource limits; a deterministic reducer $R_C$ maps terminal outputs back to the parent result. The agent chooses $C$ and supplies $x$. It neither chooses internal nodes nor observes their intermediate values in hidden mode.

For repository state $s$ and an admissible schedule/fault condition $\pi$, let $O_K(C,x,s)$ denote the normalized parent observation of the canonical execution and $O_G(C,x,s,\pi)$ the observation of the candidate. Observations include the parent output or visible failure, effects, budget accounting, and visibility-relevant metadata. The contract defines a conjunction of common and contract-specific obligations $\Phi_C(O_K,O_G)$. Qualification asks whether the concrete lowering satisfies these obligations over a finite domain $D$ and tested conditions $\Pi$:
\begin{equation}
  \forall (x,s)\in D,\ \forall \pi\in\Pi:\quad \Phi_C\!\left(O_K(C,x,s),O_G(C,x,s,\pi)\right).
  \label{eq:qualification}
\end{equation}
Equation~\ref{eq:qualification} defines finite qualification over an explicit domain and tested conditions. It is intentionally more permissive than byte equality: a contract may admit different artifact identifiers, execution accounting, semantically unordered evidence, or bounded presentation choices while requiring agreement on status, grounding, effects, and failure semantics.

\paragraph{Threat model.}
The target is accidental semantic drift introduced by the lowering or runtime: omitted branches, wrong bindings, completion-order dependence, suppressed checks, validation of the wrong state revision, timeout reclassification, provenance corruption, visibility leaks, and stale qualification evidence. The host process, hash implementation, and registry are trusted. The model excludes adversarial operating-system compromise, irreversible external side effects beyond declared modes, arbitrary agent-generated graphs, and correctness of the agent policy above the parent call.

\paragraph{Contract instances.}
\localize ranks likely fault locations and returns grounded repository evidence. Its graph has 15 tool nodes: it extracts symbols and error terms, fans out search branches, ranks paths, reads high-ranked files, and reduces the evidence into a bounded \texttt{BugContext}; the measured graph has a median of seven parallel nodes. \validatepatch applies a patch to an isolated repository copy, identifies affected scope, runs tests, import checks, and static checks, and reduces them conservatively. It has five nodes and a median of two parallel nodes. The contracts stress different obligations: localization emphasizes completeness, grounding, provenance, and ordering; patch validation emphasizes mandatory checks, conservative pass semantics, and isolation.

\section{RefinementGuard}
\label{sec:design}

\subsection{Execution substrate}

Compilation resolves the parent contract, an admissible graphlet, concrete tool implementations, and the reducer before instantiating a graph. Structural verification checks graph identity, bindings, type compatibility, acyclicity, reachability, reducer terminal coverage, declared effects, resource feasibility, and budgets. Execution uses a single-assignment blackboard and launches nodes when their inputs become available. Named preconditions run before a node; output schemas, postconditions, commit rules, effects, and timeout semantics are checked afterward. Exceptions, timeouts, cancellation, and malformed or partial outputs are normalized into explicit node evidence. The reducer consumes declared terminal outputs and cannot use completion order as semantic input.

These mechanisms prevent malformed execution but do not establish parent semantics. The evidence-separated ladder gives each layer a distinct object: \bzero checks parent shape, \bone checks the graph, \btwo checks executed nodes, and \bthree compares paired parent observations. The admission guard checks certificate applicability; Appendix Table~\ref{tab:validator-ladder} summarizes the evidence boundaries.

\subsection{Parent-semantic obligations}

\bthree evaluates named obligations rather than an opaque output-distance metric. Let $A_C$ contain every active applicable obligation for contract $C$, including a configured quality relation, and let $R_C\subseteq A_C$ contain the required non-quality obligations. The implemented decision is
\begin{equation}
\operatorname{B3}(C)=
\begin{cases}
\textsc{reject} & \exists q\in A_C:\ q \text{ conclusively fails},\\
\textsc{accept} & R_C\neq\varnothing\ \land\ \forall r\in R_C:\ r \text{ conclusively passes}\\
& \land\ \nexists q\in A_C:\ q \text{ conclusively fails},\\
\textsc{inconclusive} & \text{otherwise.}
\end{cases}
\label{eq:b3-decision}
\end{equation}
A conclusive failure of an active quality relation therefore rejects, while an inconclusive optional quality result does not block acceptance once all required obligations pass. Missing required evidence never becomes acceptance. Common predicates cover status, schema, budget, effects, visibility, failure, and schedule. Localization additionally requires applicable-branch execution, repository-grounded snippets, successful-node provenance at the input revision, canonical path order, and configured retrieval quality. Patch validation requires conservative pass, all mandatory checks, patched-revision isolation, failure preservation, and bounded visibility. Appendix~\ref{app:obligations} specifies every executable rule.

The separation is consequential. A missing localization branch produces a serializable parent result (\bzero), leaves a well-formed graph (\bone), and does not violate the contracts of nodes that still execute (\btwo). \bthree rejects it because the paired observation lacks evidence from a mandatory applicable branch. Conversely, a malformed binding is rejected by \bone and never enters the semantic denominator. The guard rejects a stale reducer digest before semantic evaluation.

\subsection{Paired qualification}

The harness copies the input repository into independent candidate and canonical snapshots and confirms their initial equality. Execution order is randomized. The canonical implementation is separately invoked and is not callable from the candidate reducer; localization scoring constants are independently declared, although their present numeric values coincide. Candidate and canonical outputs are projected into a semantic namespace before obligation evaluation, while artifact identifiers and admission metadata remain separate. Expected mutation labels are withheld from \bthree.

Qualification first establishes whether a perturbation activated: the mutation must produce an observed graph, node-evidence, parent-output, failure, effect, budget, or visibility difference. Matching output alone does not establish equivalence. Activated survivors pass an equivalence screen based on operator semantics and domain evidence; remaining ambiguous cases are adjudicated. Rejected cases are minimized and replayed, retaining their reason codes. This pipeline makes the mutation corpus a red team for the validators rather than a definition of invalidity.

\subsection{Qualification certificates and fail-closed admission}

A qualified instance produces an evidence manifest and graph-instance \qcert binding the parent contract, canonical implementation, lowering, instantiated graph, reducer, tools, executor, normalizer, observation projection, obligations, validation manifest, qualification corpus and results, mutation score and residual classes, environment, effect and resource modes, admission domain, and reproducibility digests. It attests that this exact instance passed the disclosed finite protocol; it is neither a universal correctness claim nor authorization for a graph family.

At runtime, the guard retrieves the \qcert and checks every bound artifact plus predicates over environment, requested modes, and input partition. Admission codes distinguish missing qualification, missing or invalid digest, stale artifact, environment mismatch, effect mismatch, resource mismatch, outside-domain input, and quarantine. The guard runs before graph scheduling. On rejection, the runtime records an admission artifact with \texttt{candidateStarted=false} and invokes the independent canonical implementation when available; without fallback, the parent call fails explicitly. Qualification and admission reason codes remain disjoint.

\section{Experimental method}
\label{sec:method}

\paragraph{Mutation corpora.}
The evaluation contains parent-semantic and local-contract operator families for both contracts. Semantic operators omit search branches, narrow scope, reorder or truncate localization evidence, skip reads, suppress or weaken checks, misattribute status, validate the unpatched state, and expose internal observations. Local operators violate runtime output schemas or types, postconditions, commit rules, effect declarations, and timeout/failure handling. Case selection is frozen by contract-input stratum. Of 112 generated development/qualification parent-semantic mutants, 111 activate and one is uncovered. One human adjudicator---the repository owner---reviewed every generated row: the 62 activated \bthree survivors, the uncovered row, and the 49 \bthree detections. The adjudicator could see mutation/operator identity, the expected obligation, \bzero--\bthree outcomes, output-difference summaries, candidate--reference diffs, and linked evidence. No decision received an independent second review; with one adjudicator, disagreement could not be observed and no inter-rater agreement statistic can be reported. The automated \bthree decision itself remained mutation-label blind.

The adjudication excludes 11 domain-equivalent activated cases, leaving 100 non-equivalent cases (56 localization, 44 patch): 49 \bthree rejections and 51 residual misses. Matching output alone never establishes equivalence; the 11 exclusions use domain-reachability arguments. This single-adjudicator labeling is reported as a source of uncertainty rather than folded into validator evidence.

\paragraph{Denominators and leakage control.}
Layer-specific rates use activated, non-equivalent mutant instances that reach the layer; intersections additionally report unique and cumulative detections. Local-contract mutations use a separate denominator, correct graphlets measure valid-variation acceptance, and qualification invalidations are admission cases rather than semantic mutations. Each contract's 100 controlled inputs were assigned by seeded split to 30 development, 30 qualification, and 40 held-out cases. The operator registry, mutant identities, thresholds, commands, and held-out IDs were frozen before held-out content was inspected.

The same frozen mutant instances are evaluated on qualification and held-out inputs; therefore, identical per-operator denominators measure new-input replication, not unseen-operator generalization. Here, ``frozen'' applies to mutant identities rather than an input-independent graph: \bone checks the concrete graph instantiated for each tested input. Thus a fixed mutant can pass \bone on qualification inputs and fail it on held-out inputs; such failures are structural rejects, not paired \btwo--\bthree semantic outcomes. A mutant is one denominator unit after aggregation across its input pool (15 selected qualification cases or all 40 held-out cases). In the held-out-input evaluation, 77 activated non-equivalent semantic mutants enter the ladder. Twenty-one patch mutants fail \bone structurally, leaving exactly 56 localization mutants for the paired \btwo--\bthree semantic evaluation. Equivalence exclusions reuse human decisions for the same frozen mutant identities; held-out outputs were not adjudicated separately.

\paragraph{Attribution and replay.}
A \bthree detection contributes to the separation result only if an obligation relevant to the mutated behavior fails, its reason code reproduces, and the evidence arises from the paired semantic observations rather than a mutation label or admission mismatch. Counterexamples are stored with candidate and reference observations, obligation results, graph/evidence references, and replay reports. The full replay corpus contains 2,768 reason-code instances.

\paragraph{Performance and quality.}
E5 compares serial reference, concurrency-disabled candidate, visible graph, handcrafted parallel, unguarded hidden, and guarded hidden execution. Forty held-out inputs per contract run in zero-delay and controlled asynchronous-delay regimes. The frozen controlled configuration declares these base delays in milliseconds: default and \texttt{listFiles}, 25; \texttt{grep}, 80; \texttt{readFile}, 50; \texttt{parseTraceback}, 15; and \texttt{findSymbols}, 40. In the evaluated graphlets, injected asynchronous waits are exercised by localization's \texttt{grep} and \texttt{readFile} calls and patch validation's \texttt{runTargetedTests} calls (configured under \texttt{runTest}), which use the 25-ms default; the other named entries are inactive. Each exercised delay equals its base times a seed-derived uniform factor in $[0.9,1.1]$. The same factor is deterministically reused within a repetition. Repetition seeds 1729, 2718, and 3141 are shared across systems, pairing the injected delays; system order is randomized with seed 20260803. The zero-delay regime sets every base and jitter to zero. E5 runs one aggregate-excluded warmup for each contract--regime--system configuration; each case--regime--system cell then has three measured repetitions, yielding 24 warmups and 2,880 measured executions with no failures. E6 derives quality metrics from persisted E5 outputs. A separate 10,000-iteration microbenchmark measures only admission.

\paragraph{Robustness.}
Schedule tests exercise FIFO, criticality, concurrency-one, maximum-concurrency, and randomized policies. Fault tests inject stale or delayed responses, subprocess failures, corrupt provenance, exceptions, timeouts, empty or partial results, malformed or duplicate outputs, reducer exceptions, and cancellation. Schedule coverage is policy-based rather than exhaustive enumeration of topological orders. Patch schedule/fault configurations fail at \bone and therefore produce no \btwo/\bthree robustness evidence; robustness beyond \bone is a localization case study.

\paragraph{Statistics.}
Mutation sensitivity uses an instance-weighted cluster bootstrap over contract--operator families (10,000 resamples; seed 20260815), with operator macro-averages and per-family rates. Wilson intervals remain descriptive for unclustered layer counts. Paired binary comparisons use exact two-sided McNemar tests. For latency, the three repetitions are first collapsed to a median within each case and system; paired inference then bootstraps the 40 cases (10,000 resamples; seed 20260815). The \bthree ablation and equality comparison reaggregate frozen observations offline.

\section{Results}
\label{sec:results}

\subsection{Q1: Do lower layers miss semantic defects?}

The control population confirms that \btwo measures a distinct layer: all 48 activated local-contract defects pass \bone and all 48 are rejected by \btwo. All 60 correct graphlets pass the pipeline.

\begin{figure}[t]
  \centering
  \includegraphics[width=\linewidth]{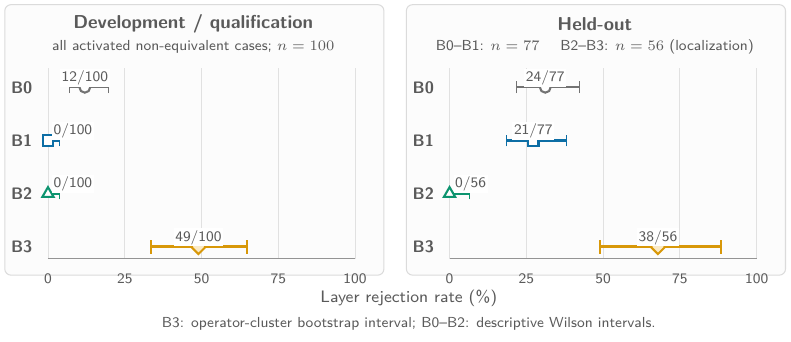}
  \caption{Parent-semantic defect rejection by layer. \bthree whiskers use contract--operator cluster bootstrap intervals; \bzero--\btwo use descriptive Wilson intervals. Held-out-input \bzero/\bone use 77 mutants, whereas \btwo/\bthree use the 56 \bone-passing localization mutants.}
  \label{fig:separation}
\end{figure}

On the 100 non-equivalent development mutations, \bzero rejects 12 and \bthree rejects 49; \bone and \btwo reject none (Figure~\ref{fig:separation}). The full intersection is more informative than marginal rates: all 12 \bzero rejections are also rejected by \bthree, while 37 additional \bthree detections pass \bzero--\btwo. Thus \bthree rejects 37/88 (42.0\%) of cases accepted by every lower layer, and 37/49 (75.5\%) of its detections are unique beyond that ladder. Cumulative \bzero--\bthree detection is 49/100. The instance-weighted operator-cluster bootstrap interval for \bthree is [33.7\%, 64.8\%], and the operator macro-average is 50.1\% across 26 contract--operator families. Per-family results appear in Appendix~\ref{app:clustered}.

The disjoint held-out-input evaluation contains 77 activated parent-semantic mutants. Twenty-one patch mutants fail \bzero and \bone structurally, so the held-out \btwo--\bthree semantic result is localization-only. Among 56 localization mutants, three are jointly rejected by \bzero and \bthree, 35 pass \bzero--\btwo and are uniquely rejected by \bthree, and 18 pass all layers. Consequently, the cumulative ladder rejects 59/77, while \bthree rejects 35/53 cases accepted by every lower layer. Its localization semantic rejection rate is 38/56 (67.9\%; operator-cluster CI [49.1\%, 88.5\%]) across the same 12 frozen operator families. The identical family denominators in Appendix~\ref{app:clustered} are by design: the mutant instances are fixed while the input pool changes. The paired \btwo--\bthree comparisons remain decisive in development/qualification ($n_{01}=49,n_{10}=0,p=3.55\times10^{-15}$) and on held-out localization inputs ($38,0,p=7.28\times10^{-12}$).

The separation is operationally meaningful. Omitting an applicable localization search branch leaves executed nodes locally valid but fails branch completeness. Checking an unpatched snapshot can preserve every node schema yet fail patch isolation. Invalid producer bindings are rejected structurally and do not inflate the semantic-only count.

\subsection{Q2: Does B3 balance sensitivity and valid variation?}

The objective is selective semantic admission, not maximum mutation rejection. All 49 development/qualification \bthree detections have attributable obligation evidence, and all 2,768 stored reason-code instances replay. Most failures concern localization ordering, schema refinement, branch completeness, patch sandbox isolation, localization quality, and visibility; counts overlap.

\begin{figure}[t]
  \centering
  \includegraphics[width=\linewidth]{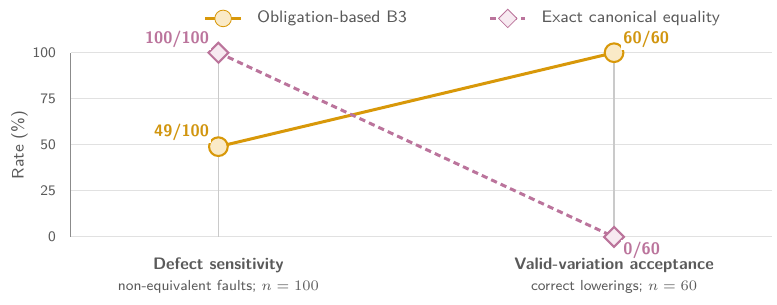}
  \caption{The central semantic tradeoff on separate fault and valid-implementation populations. Exact equality rejects 100/100 faults but also all 60 valid candidates (30 per contract); \bthree accepts 60/60 valid variants while retaining attributable 49/100 fault sensitivity.}
  \label{fig:attribution-ablation}
\end{figure}

Figure~\ref{fig:attribution-ablation} makes the tradeoff explicit. Equality has higher observed fault sensitivity (100/100 versus 49/100), but it rejects every correct qualification graphlet: 0/30 localization and 0/30 patch implementations are accepted. Valid differences include execution accounting, artifact/provenance identity, floating-point representation, and contract-permitted evidence presentation. \bthree therefore accepts 60/60 valid lowerings while rejecting and explaining 49/100 faults.

The 51 residual misses make the boundary of the current specification concrete: 45 alter task-semantic fields and six alter accounting or visibility. Frequent families include ranking/confidence perturbations, truncation-boundary evidence substitutions, check-status changes, and weakened validations within current thresholds. Retaining and releasing these counterexamples creates a direct refinement queue for the next obligation version.

\subsection{Q3: Does hidden execution retain its benefits?}

\begin{figure}[t]
  \centering
  \includegraphics[width=\linewidth]{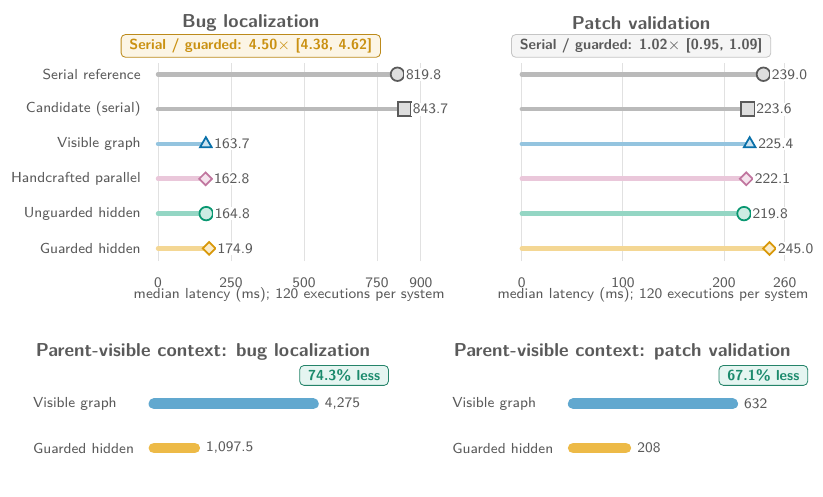}
  \caption{Controlled-latency speed and parent-visible context. The 4.50$\times$ callout is paired over 40 localization inputs; localization has parallel slack, whereas patch validation does not.}
  \label{fig:performance}
\end{figure}

The admission check averages 0.00693\,ms for localization and 0.00744\,ms for patch validation (10,000 iterations). In the controlled asynchronous-delay regime specified in Section~\ref{sec:method}, localization medians are 819.8\,ms serial, 164.8\,ms unguarded hidden, and 174.9\,ms guarded hidden (Figure~\ref{fig:performance}). Across 40 inputs, after taking within-case medians of three repetitions, the paired serial/guarded latency ratio is 4.50$\times$ (case-bootstrap CI [4.38, 4.62]); guarded-minus-serial latency is $-598.2$\,ms [${-606.8}$, ${-593.2}$]. The guard adds 13.9\,ms [8.4, 20.3] relative to unguarded hidden execution.

The 4.50$\times$ result is regime- and contract-specific. Patch validation shows no demonstrated speedup: its serial/guarded ratio is 1.02$\times$ [0.95, 1.09]. With zero injected delay, guarded localization takes 53.5\,ms versus 30.5\,ms serial, and guarded patch validation takes 255.7\,ms versus 244.6\,ms serial. These results expose orchestration overhead when external work is cheap and identify the deployment criterion: semantic qualification controls correctness, while measured parallel slack determines whether the qualified lowering is worthwhile.

Hidden execution reduces median parent-visible context from 4,275 to 1,097.5 tokens for localization (74.3\%) and from 632 to 208 for patch validation (67.1\%). E6 observes no quality degradation on the controlled workload: all 24 contract--system--delay tolerance cells pass; localization MRR and recall metrics remain 1.0, and patch false-pass and false-fail rates remain zero.

\section{Related work}
\label{sec:related}

\paragraph{Agent harnesses and orchestration.}
ReAct, Toolformer, and SWE-agent establish tool use and interface design as components of agent behavior \citep{yao2023react,schick2023toolformer,yang2024sweagent}. LLMCompiler parallelizes function calls \citep{kim2024llmcompiler}, and the code-as-harness view treats execution, state, tools, and verification as first-class agent infrastructure \citep{ning2026codeharness}. \system instead validates a hidden implementation substituted beneath an already selected tool; dataflow scheduling itself is conventional \citep{dean2004mapreduce,isard2007dryad}.

\paragraph{Agent verification and runtime enforcement.}
ToolEmu and AgentDojo evaluate risks in tool environments \citep{ruan2024toolemu,debenedetti2024agentdojo}. AgentSpec and Agent-C enforce proposed-action policies \citep{wang2026agentspec,kamath2025agentc}; concurrent preprints study probabilistic unsafe-state prediction, behavioral policy, verified workflow models, and canonical action identity \citep{wang2025probguard,bhardwaj2026abc,miculicich2025veriguard,wang2026lean4agent,wang2026cava}. These systems verify agent actions, policies, or trajectories. \system targets a different object: whether an effectful hidden lowering preserves the parent operation already selected by the agent. Table~\ref{tab:positioning} gives the detailed comparison.

\paragraph{Contracts and translation validation.}
Design by Contract supplies local pre/postconditions \citep{meyer1992contract}; translation validation checks transformation results \citep{pnueli1998translation,lopes2021alive2}, while CompCert mechanizes compiler correctness \citep{leroy2009compcert}. The novelty here is translation validation across an effectful agent--tool boundary, followed by admission bound to the tested artifacts and domain. The \qcert borrows proof-carrying code's producer/consumer split \citep{necula1997pcc}, but carries finite, not deductive, evidence.

\paragraph{Mutation and counterexamples.}
Mutation testing measures validator sensitivity \citep{jia2011mutation}, while counterexample-guided methods refine specifications \citep{solarlezama2006sketching}. Here, mutations probe sensitivity, valid variants measure tolerance, and residuals drive obligation evolution.

\section{Discussion and scope}
\label{sec:limitations}

\system applies when an agent-facing contract has an optimized harness and independent fallback. Localization realizes speed and context gains under controlled latency; patch validation shows that qualification and optimization selection remain separate decisions.

The evidence covers two repository-level contracts on controlled repositories. The disjoint-input replication confirms localization behavior for frozen operators, not unseen-operator generalization; 21 patch mutants fail structurally before the held-out \btwo--\bthree comparison. The 51 residuals form a concrete obligation-refinement queue alongside the paper's attributable detections.

Human adjudication used a single reviewer with access to mutation identity, expected obligations, layer outcomes, candidate--reference diffs, and linked evidence; inter-rater agreement was therefore not estimated. The automated \bthree decision remained mutation-label blind. Qualification is specific to the evaluated graph instance and disclosed domain, and relevant changes require requalification.

Preventing silent semantic drift in hidden runtimes can improve the reliability of agent deployments, but treating finite qualification or local-copy isolation as a broader guarantee could create misplaced confidence in systems operating outside the evaluated conditions.

Future work should add structurally valid negative patch cases, unseen operators, remote effects, real repositories, live-model trajectories, and independent adjudication.
\section{Conclusion}

\system validates hidden, effectful agent--tool lowerings against an independent reference and binds the finite result to fail-closed admission. Across two repository contracts, \bthree contributes 37 unique detections among 88 lower-layer survivors and accepts all 60 valid variants rejected by equality. Under controlled tool latency, localization retains a 4.50$\times$ serial/guarded median-latency ratio and reduced parent-visible context; patch validation shows that qualification need not imply faster execution. Hidden optimization should carry independent evidence that the admitted harness still implements the selected operation.

\bibliographystyle{plainnat}
{\small
\bibliography{references}
}

\clearpage
\appendix
\section*{Appendix}

\section{Formal scope and implementation}
\label{app:formal}

\paragraph{Finite qualification object.}
A graph-instance \qcert records references to the contract, canonical implementation, lowering, graph, reducer, tools, runtime, normalizer, projection, obligations, and validation manifest. Each reference includes identity, version, digest, and, where retained, an artifact path. It also stores the validation domain, guard expression, assurance level, case/schedule/fault counts, mutation score, residual-class summary, obligation summary, environment, modes, and reproducibility digest. Runtime admission recomputes these bindings over the current call. The object attests to completion of that disclosed protocol for one instance; it does not authorize future graph variants.

\paragraph{Executor invariants.}
The concrete runtime enforces acyclicity, reachability, complete producer bindings, compatible schemas and types, reducer terminal coverage, effect and resource declarations, single-assignment blackboard state, node preconditions and postconditions, commit rules, and timeout/failure normalization. Value-ready scheduling supports FIFO, criticality, randomized, and concurrency-limited policies. Sanitized evidence retains the facts needed by local validation while excluding arbitrary raw runtime state.

\begin{table}[ht]
\caption{Contract instances used in the evaluation. Parallel-node counts are medians over measured graph executions.}
\label{tab:contracts}
\centering
\small
\begin{tabularx}{\linewidth}{@{}lrrY@{}}
\toprule
Contract & Nodes & Parallel nodes & Principal semantic obligations \\
\midrule
\localize & 15 & 7 & Applicable branch completeness; grounded evidence; provenance; allowed ordering; localization quality; visibility \\
\validatepatch & 5 & 2 & Conservative pass; mandatory checks; patched-state revision; failure preservation; visibility \\
\bottomrule
\end{tabularx}
\end{table}

\begin{table}[ht]
\caption{Validator and guard evidence boundaries. Each layer addresses a distinct failure class; the admission guard is not a semantic validator.}
\label{tab:validator-ladder}
\centering
\small
\begin{tabularx}{\linewidth}{@{}lYYY@{}}
\toprule
Layer & Available evidence & Principal checks & Deliberate blind spot \\
\midrule
\bzero & Contract, candidate parent outcome & Schema, serialization, failure envelope, budgets & Graph and reference behavior \\
\bone & Graph, bindings, declarations & Topology, types, reachability, effects, resources & Concrete execution and parent meaning \\
\btwo & Graph, local contracts, node evidence & Node schema, pre/postconditions, commit, effect, timeout & Reference parent observation \\
\bthree & Paired parent observations, obligations & Parent-visible semantic refinement & Qualification admission and mutation label \\
Guard & \qcert, current bindings, call context & Artifact, environment, mode, and domain match & Semantic re-evaluation \\
\bottomrule
\end{tabularx}
\end{table}

\begin{table}[ht]
\caption{Positioning by verified object and enforcement point. Citations appear in each row so the comparison is self-contained.}
\label{tab:positioning}
\centering
\footnotesize
\begin{tabularx}{\linewidth}{@{}lYY@{}}
\toprule
Work & Verified object / method & \system boundary \\
\midrule
AgentSpec / Agent-C \citep{wang2026agentspec,kamath2025agentc} & Proposed actions; rules or SMT & Below an already selected tool \\
ABC / VeriGuard \citep{bhardwaj2026abc,miculicich2025veriguard} & Behavioral or safety policy & Harness translation correctness \\
ProbGuard \citep{wang2025probguard} & Unsafe-state prediction; probabilistic monitoring & Paired parent observations \\
Lean4Agent \citep{wang2026lean4agent} & Workflow/trajectory; Lean proofs & Concrete hidden lowering \\
CAVA \citep{wang2026cava} & Canonical action identity; attestation & Parent-operation refinement \\
\system & Hidden effectful lowering; paired obligations & Artifact-bound admission \\
\bottomrule
\end{tabularx}
\end{table}

\section{Operational obligation specification}
\label{app:obligations}

Table~\ref{tab:obligations-full} specifies the predicates implemented by the frozen \bthree validator. ``Candidate'' and ``canonical'' refer to normalized parent observations; required evidence that is absent or nonconclusive yields \textsc{inconclusive}, never \textsc{accept}.

\begin{table}[ht]
\caption{Frozen parent-semantic obligations and conclusive decision rules.}
\label{tab:obligations-full}
\centering
\scriptsize
\begin{tabularx}{\linewidth}{@{}lY@{}}
\toprule
Obligation & Conclusive rule \\
\midrule
Status refinement & Candidate success requires canonical success. \\
Schema refinement & On candidate success, validate output against the registered parent schema. \\
Budget refinement & Candidate output object-token count does not exceed the configured parent budget. \\
Effect refinement & Source-root digest is unchanged and declared parent-visible effects exactly equal canonical effects. \\
Visibility refinement & No runtime value IDs, sandbox paths, fault controls, or forbidden raw traces escape. \\
Failure refinement & Canonical visible failure is not converted to candidate success. \\
Schedule refinement & With a FIFO baseline, normalized parent observations agree exactly; otherwise not applicable. \\
Localization branch completeness & Every required branch ID is nonempty, activated, and successful. \\
Localization grounding & Every candidate path exists; every snippet text and line interval matches repository content. \\
Localization provenance & Every snippet records operation, node, path, revision, and interval; node succeeded and revision equals the input-root digest. \\
Localization ordering & Candidate \texttt{candidateFiles} paths exactly preserve canonical ranked path order. \\
Localization quality & Configured MRR threshold and, when enabled, recall@5 hold against frozen ground truth. \\
Patch conservative pass & Candidate pass requires canonical pass; conservative candidate failure is permitted. \\
Patch mandatory checks & Tests, import, and static checks are present, activated, and successful. \\
Patch sandbox isolation & Source is unchanged; checks bind the patched digest; all checks are recorded; sandbox is disposed. \\
Patch failure preservation & Declared failed nodes and canonical failed tests/statuses are retained; candidate cannot pass with a failed node. \\
Patch visibility & No sandbox paths, unbounded stdout/stderr, raw traces, or fault-control state escape. \\
\bottomrule
\end{tabularx}
\end{table}

\section{Mutation accounting and denominators}
\label{app:mutation}

Table~\ref{tab:rq1-full} makes the denominator transitions explicit. Structural rates for the held-out-input corpus use 77 activated, non-equivalent parent-semantic mutants. The paired semantic rate uses only the 56 mutants that pass \bone; these are all localization mutants. Local-contract mutations and correct implementations use separate evaluation units.

\begin{table}[ht]
\caption{Validator separation with exact numerators, denominators, and Wilson 95\% confidence intervals.}
\label{tab:rq1-full}
\centering
\small
\begin{tabular}{@{}llrrl@{}}
\toprule
Corpus / family & Layer & Rejected & Total & Wilson 95\% CI \\
\midrule
Local-contract, activated & \bone & 0 & 48 & [0.0, 7.4]\% \\
Local-contract, activated & \btwo & 48 & 48 & [92.6, 100.0]\% \\
\midrule
Development semantic & \bzero & 12 & 100 & [7.0, 19.8]\% \\
Development semantic & \bone & 0 & 100 & [0.0, 3.7]\% \\
Development semantic & \btwo & 0 & 100 & [0.0, 3.7]\% \\
Development semantic & \bthree & 49 & 100 & [39.4, 58.7]\% \\
\midrule
Held-out-input, structural & \bzero & 24 & 77 & [21.9, 42.2]\% \\
Held-out-input, structural & \bone & 21 & 77 & [18.6, 38.1]\% \\
Held-out-input, B1-passing localization & \btwo & 0 & 56 & [0.0, 6.4]\% \\
Held-out-input, B1-passing localization & \bthree & 38 & 56 & [54.8, 78.6]\% \\
\bottomrule
\end{tabular}
\end{table}

The development/qualification corpus generates 112 parent-semantic mutants. Of these, 111 activate and one is \texttt{UNCOVERED}; the uncovered case is outside every rejection-rate denominator. The human adjudicator reviewed the 62 activated \bthree survivors, the uncovered row, and the 49 \bthree detections. The review identifies 11 domain-equivalent cases, leaving exactly 100 non-equivalent cases: 56 localization and 44 patch. \bthree rejects 49 (33 localization, 16 patch), and the remaining 51 are \texttt{REAL\_MISS}. No decision was independently reviewed; disagreement was therefore unobservable and no inter-rater statistic is available. Sixty unmutated graphlet cases pass.

The held-out protocol freezes the same mutation registry and mutant identities but changes the inputs. Each contract's 100 controlled inputs are split by seed into 30 development, 30 qualification, and 40 held-out cases. A mutant contributes one denominator unit after aggregation over 15 selected qualification inputs or all 40 held-out inputs. Thus the repeated localization denominators in Table~\ref{tab:operator-rates} are expected and do not indicate reuse of the same input executions. The result measures input-case replication within the same generator and operator set, not generalization to unseen operators. Twenty-one held-out patch mutants fail \bone, so the held-out \btwo--\bthree semantic denominator contains 56 localization mutants only.

\begin{table}[ht]
\caption{Complete parent-semantic layer intersections. ``Unique B3'' passes \bzero--\btwo.}
\label{tab:intersections}
\centering
\small
\begin{tabular}{@{}lrrrrr@{}}
\toprule
Corpus & Lower only & Lower + B3 & Unique B3 & All accept & Cumulative reject \\
\midrule
Development ($n=100$) & 0 & 12 & 37 & 51 & 49 (49.0\%) \\
Held-out inputs ($n=77$) & 21 & 3 & 35 & 18 & 59 (76.6\%) \\
\bottomrule
\end{tabular}
\end{table}

Conditioning on passage through all lower layers, \bthree rejects 37/88 development/qualification mutants (42.0\%) and 35/53 held-out-input localization mutants (66.0\%). In development/qualification, all 12 \bzero detections overlap \bthree; on held-out inputs, the 21 lower-only mutants are patch structural rejects and the three overlap mutants are localization mutations. Exact paired comparisons follow.

\begin{table}[ht]
\caption{Exact paired \btwo--\bthree comparisons on B1-passing non-equivalent semantic cases. $n_{01}$ counts \btwo accept/\bthree reject; $n_{10}$ counts the reverse.}
\label{tab:mcnemar}
\centering
\small
\begin{tabular}{@{}lrrrr@{}}
\toprule
Corpus & Paired $n$ & $n_{01}$ & $n_{10}$ & Exact two-sided $p$ \\
\midrule
Development / qualification & 100 & 49 & 0 & $3.55\times10^{-15}$ \\
Held-out-input localization & 56 & 38 & 0 & $7.28\times10^{-12}$ \\
\bottomrule
\end{tabular}
\end{table}

\section{Attribution, reason codes, and misses}
\label{app:attribution}

\begin{table}[ht]
\caption{Stored reason-code instances for development \bthree detections. Counts are nonexclusive and sum to 58 across 49 rejected cases.}
\label{tab:reasons}
\centering
\small
\begin{tabular}{@{}lr@{\qquad}lr@{}}
\toprule
Reason code & Count & Reason code & Count \\
\midrule
\texttt{OUTPUT\_SEMANTIC\_MISMATCH} & 23 & \texttt{VISIBILITY\_LEAK} & 6 \\
\texttt{OUTPUT\_SCHEMA\_MISMATCH} & 10 & \texttt{WRONG\_STATE\_REVISION} & 5 \\
\texttt{MANDATORY\_BRANCH\_OMITTED} & 9 & \texttt{PROVENANCE\_INVALID} & 2 \\
\texttt{BUDGET\_VIOLATION} & 2 & \texttt{PROVENANCE\_MISSING} & 1 \\
\bottomrule
\end{tabular}
\end{table}

The 51 real misses contain 45 task-semantic field changes and six accounting or visibility changes. Frequent mechanisms include ranking or confidence perturbation, alternate evidence near a truncation boundary, dropped or misattributed check status, narrowed visibility, and weakened validation checks that remain inside present thresholds. They are retained in the released adjudication worksheet and are not included among \bthree detections.

\section{Exploratory B3 ablation}
\label{app:ablation}

The ablation reaggregates frozen per-case obligation results and parent observations. It does not rerun systems, modify the production validator, or change issued \qcert{}s. ``Without G/P/V'' removes localization grounding, localization provenance, common visibility refinement, and patch visibility. Equality-only rejects whenever normalized parent-observation content digests differ and ignores $\Phi_C$.

\begin{table}[ht]
\caption{Exploratory B3 ablation. The $p$ columns report exact McNemar tests versus full \bthree. Equality-only is a diagnostic comparator, not an attribution-preserving validator.}
\label{tab:ablation}
\centering
\small
\begin{tabular}{@{}lrrlrrl@{}}
\toprule
& \multicolumn{3}{c}{Development / qualification} & \multicolumn{3}{c}{Held-out-input localization} \\
\cmidrule(lr){2-4}\cmidrule(l){5-7}
Variant & Reject & Total & $p$ vs. full & Reject & Total & $p$ vs. full \\
\midrule
Full B3 & 49 & 100 & -- & 38 & 56 & -- \\
Common only & 16 & 100 & $2.33\times10^{-10}$ & 5 & 56 & $2.33\times10^{-10}$ \\
Without G/P/V & 44 & 100 & 0.0625 & 36 & 56 & 0.5 \\
Without provenance & 49 & 100 & 1.0 & 38 & 56 & 1.0 \\
Canonical equality only & 100 & 100 & $8.88\times10^{-16}$ & 56 & 56 & $7.63\times10^{-6}$ \\
\bottomrule
\end{tabular}
\end{table}

Common obligations alone lose 33 of 49 development detections and 33 of 38 held-out detections, showing that contract-specific obligations account for most incremental sensitivity. Provenance has no unique detection because every provenance failure is co-detected. On the separate correct-implementation population, full \bthree accepts 60/60 while equality accepts 0/60 (30 localization and 30 patch). Equality's additional mutation sensitivity therefore comes with a 100\% false-rejection rate on these valid variants.

\section{Clustered mutation inference}
\label{app:clustered}

The inferential unit is a contract--mutation-operator cluster. Resampling all instances belonging to selected clusters preserves within-family dependence. Development/qualification has 26 clusters and an instance-weighted rate of 49/100 with cluster-bootstrap 95\% CI [33.7\%, 64.8\%]; the operator macro-average is 50.1\% [34.7\%, 65.9\%]. Held-out-input localization has 12 clusters and a rate of 38/56 with interval [49.1\%, 88.5\%]; the macro-average is 73.8\% [55.4\%, 90.3\%]. Table~\ref{tab:operator-rates} reports every cluster; repeated denominators reflect the frozen mutant set, while numerators are recomputed on the disjoint input pool.

\begin{table}[ht]
\caption{\bthree rejection by contract--operator cluster. Held-out-input rows reuse the frozen mutant instances on disjoint inputs, so matching denominators are by design.}
\label{tab:operator-rates}
\centering
\scriptsize
\begin{tabular}{@{}lllrl@{}}
\toprule
Corpus & Contract & Operator & Reject & Rate \\
\midrule
Dev. & \localize & \nolinkurl{completion_order_ranking} & 3/3 & 100\% \\
Dev. & \localize & \nolinkurl{drop_error_search_branch} & 3/3 & 100\% \\
Dev. & \localize & \nolinkurl{drop_provenance} & 3/3 & 100\% \\
Dev. & \localize & \nolinkurl{drop_symbol_search_branch} & 4/4 & 100\% \\
Dev. & \localize & \nolinkurl{leak_raw_observation} & 2/3 & 66.7\% \\
Dev. & \localize & \nolinkurl{narrow_search_scope} & 2/7 & 28.6\% \\
Dev. & \localize & \nolinkurl{omit_rank_input} & 0/6 & 0\% \\
Dev. & \localize & \nolinkurl{read_wrong_candidate_index} & 4/5 & 80\% \\
Dev. & \localize & \nolinkurl{skip_candidate_read} & 3/5 & 60\% \\
Dev. & \localize & \nolinkurl{swap_search_term_source} & 4/6 & 66.7\% \\
Dev. & \localize & \nolinkurl{truncate_candidate_read} & 3/5 & 60\% \\
Dev. & \localize & \nolinkurl{truncate_localization_output} & 2/6 & 33.3\% \\
Dev. & \validatepatch & \nolinkurl{drop_check_status} & 0/3 & 0\% \\
Dev. & \validatepatch & \nolinkurl{drop_patch_diagnostics} & 6/6 & 100\% \\
Dev. & \validatepatch & \nolinkurl{expose_patch_internal_state} & 1/4 & 25\% \\
Dev. & \validatepatch & \nolinkurl{inject_patch_failure} & 0/4 & 0\% \\
Dev. & \validatepatch & \nolinkurl{leak_full_subprocess_output} & 3/3 & 100\% \\
Dev. & \validatepatch & \nolinkurl{misattribute_check_status} & 0/3 & 0\% \\
Dev. & \validatepatch & \nolinkurl{misreport_check_status} & 0/3 & 0\% \\
Dev. & \validatepatch & \nolinkurl{shorten_check_timeout} & 0/1 & 0\% \\
Dev. & \validatepatch & \nolinkurl{suppress_failed_tests} & 1/3 & 33.3\% \\
Dev. & \validatepatch & \nolinkurl{truncate_patch_output} & 0/5 & 0\% \\
Dev. & \validatepatch & \nolinkurl{validate_unpatched_sandbox} & 3/3 & 100\% \\
Dev. & \validatepatch & \nolinkurl{weaken_import_check} & 0/3 & 0\% \\
Dev. & \validatepatch & \nolinkurl{weaken_static_check} & 1/1 & 100\% \\
Dev. & \validatepatch & \nolinkurl{weaken_test_check} & 1/2 & 50\% \\
\midrule
Held-out inputs & \localize & \nolinkurl{completion_order_ranking} & 3/3 & 100\% \\
Held-out inputs & \localize & \nolinkurl{drop_error_search_branch} & 3/3 & 100\% \\
Held-out inputs & \localize & \nolinkurl{drop_provenance} & 3/3 & 100\% \\
Held-out inputs & \localize & \nolinkurl{drop_symbol_search_branch} & 4/4 & 100\% \\
Held-out inputs & \localize & \nolinkurl{leak_raw_observation} & 2/3 & 66.7\% \\
Held-out inputs & \localize & \nolinkurl{narrow_search_scope} & 2/7 & 28.6\% \\
Held-out inputs & \localize & \nolinkurl{omit_rank_input} & 1/6 & 16.7\% \\
Held-out inputs & \localize & \nolinkurl{read_wrong_candidate_index} & 5/5 & 100\% \\
Held-out inputs & \localize & \nolinkurl{skip_candidate_read} & 4/5 & 80\% \\
Held-out inputs & \localize & \nolinkurl{swap_search_term_source} & 6/6 & 100\% \\
Held-out inputs & \localize & \nolinkurl{truncate_candidate_read} & 3/5 & 60\% \\
Held-out inputs & \localize & \nolinkurl{truncate_localization_output} & 2/6 & 33.3\% \\
\bottomrule
\end{tabular}
\end{table}

\Needspace{14\baselineskip}
\section{Admission matrix and integration evidence}
\label{app:admission}

\begin{table}[ht]
\caption{Runtime admission decision matrix. ``Fallback'' means the canonical path is selected when available. The coverage column separates directly executed cases from source-audited branches.}
\label{tab:admission}
\centering
\scriptsize
\begin{tabular}{@{}>{\raggedright\arraybackslash}p{0.95in}>{\raggedright\arraybackslash}p{1.58in}cc>{\raggedright\arraybackslash}p{0.92in}@{}}
\toprule
Condition & Stable decision code & \shortstack{Candidate\\starts} & \shortstack{Canonical\\fallback} & Coverage \\
\midrule
All bindings and guard clauses match & \nolinkurl{execute_certified_graphlet} & yes & no & unit + E5 \\
Qualification absent & \nolinkurl{fallback_missing_certificate} & no & yes & unit + smoke + integration \\
Bound digest absent & \nolinkurl{fallback_invalid_digest} & no & yes & source audit \\
Bound digest changed & \nolinkurl{fallback_stale_artifact} & no & yes & unit + smoke \\
Environment clause fails & \nolinkurl{fallback_environment_mismatch} & no & yes & source audit \\
Effect mode differs & \nolinkurl{fallback_effect_mode} & no & yes & source audit \\
Resource mode differs & \nolinkurl{fallback_resource_mode} & no & yes & source audit \\
Input outside domain & \nolinkurl{fallback_outside_domain} & no & yes & unit + smoke \\
Qualification quarantined & \nolinkurl{fallback_quarantined} & no & yes & unit + smoke \\
Rejection and no fallback & \nolinkurl{failure_fallback_unavailable} & no & no & source audit \\
\bottomrule
\end{tabular}
\end{table}

Two graph-instance \qcert{}s were issued: localization \nolinkurl{cert-4a39fd0e677b4cfd99fbad91179330cf} and patch \nolinkurl{cert-656164e040314483aebbc01c6980a5e4}. Each has an evidence manifest and a stable reproducibility digest. Graph-family qualification was not activated.

\begin{table}[ht]
\caption{Fake-provider agent-boundary smoke. A checkmark denotes a satisfied assertion. The partial patch result is disclosed rather than used as live-agent evidence.}
\label{tab:agent-smoke}
\centering
\small
\begin{tabularx}{\linewidth}{@{}Ycc@{}}
\toprule
Check & \localize & \validatepatch \\
\midrule
Registered parent semantic tool called & \yes & \yes \\
Caller did not select internal nodes & \yes & \yes \\
Parent cannot observe hidden internal state & \yes & \yes \\
Trajectory references admission artifact & \yes & \yes \\
Stale variant prevented candidate start & \yes & \yes \\
Valid qualification admitted & \yes & \no \\
Parent received contracted result & \yes & \no \\
Canonical fallback preserved parent interface & \yes & \no \\
\bottomrule
\end{tabularx}
\end{table}

\section{Performance and quality details}
\label{app:performance}

\begin{table}[ht]
\caption{E5 controlled-delay medians over 120 measured executions per contract/system. Context is the median number of tokens exposed at the parent interface.}
\label{tab:e5-primary}
\centering
\small
\begin{tabular}{@{}lrr@{\qquad}lrr@{}}
\toprule
\multicolumn{3}{c}{\localize} & \multicolumn{3}{c}{\validatepatch} \\
\cmidrule(lr){1-3}\cmidrule(l){4-6}
System & Latency ms & Context & System & Latency ms & Context \\
\midrule
Serial canonical & 819.8 & 1205.0 & Serial canonical & 239.0 & 193 \\
No-parallel candidate & 843.7 & 1097.5 & No-parallel candidate & 223.6 & 208 \\
Visible graph & 163.7 & 4275.0 & Visible graph & 225.4 & 632 \\
Handcrafted parallel & 162.8 & 1097.5 & Handcrafted parallel & 222.1 & 208 \\
Unguarded hidden & 164.8 & 1097.5 & Unguarded hidden & 219.8 & 208 \\
Guarded hidden & 174.9 & 1097.5 & Guarded hidden & 245.0 & 208 \\
\bottomrule
\end{tabular}
\end{table}

\begin{table}[ht]
\caption{E5 zero-delay medians over 120 measured executions per contract/system. The regime isolates orchestration overhead when external tool work is cheap.}
\label{tab:e5-zero}
\centering
\small
\begin{tabular}{@{}lrr@{\qquad}lrr@{}}
\toprule
\multicolumn{3}{c}{\localize} & \multicolumn{3}{c}{\validatepatch} \\
\cmidrule(lr){1-3}\cmidrule(l){4-6}
System & Latency ms & Context & System & Latency ms & Context \\
\midrule
Serial canonical & 30.5 & 1205.0 & Serial canonical & 244.6 & 193 \\
No-parallel candidate & 27.4 & 1097.5 & No-parallel candidate & 230.9 & 208 \\
Visible graph & 24.9 & 4275.0 & Visible graph & 223.9 & 632 \\
Handcrafted parallel & 25.2 & 1097.5 & Handcrafted parallel & 231.2 & 208 \\
Unguarded hidden & 26.4 & 1097.5 & Unguarded hidden & 226.0 & 208 \\
Guarded hidden & 53.5 & 1097.5 & Guarded hidden & 255.7 & 208 \\
\bottomrule
\end{tabular}
\end{table}

\begin{table}[ht]
\caption{Case-level controlled-delay inference for guarded hidden execution. Three repetitions are collapsed within each case; differences are guarded minus comparator and intervals bootstrap 40 paired cases.}
\label{tab:e5-paired}
\centering
\small
\begin{tabular}{@{}llrrl@{}}
\toprule
Contract & Comparator & Cases & Median difference (ms) & Comparator/guarded 95\% CI \\
\midrule
\localize & Serial canonical & 40 & $-598.2$ & [4.38, 4.62] \\
\localize & Unguarded hidden & 40 & $+13.9$ & [0.89, 0.95] \\
\validatepatch & Serial canonical & 40 & $-5.6$ & [0.95, 1.09] \\
\validatepatch & Unguarded hidden & 40 & $+14.3$ & [0.84, 0.99] \\
\bottomrule
\end{tabular}
\end{table}

The case-level localization speedup estimate is 4.4988. The sign of a paired median difference need not equal the difference of separately computed group medians; Table~\ref{tab:e5-paired} reports the inferential statistic. The pure guard microbenchmark uses 10,000 iterations per contract. Mean live-recomputed-digest time is 0.00693\,ms for localization and 0.00744\,ms for patch; mean embedded-digest time is 0.00770 and 0.00720\,ms, respectively.

\begin{table}[ht]
\caption{E6 quality summary. Metrics are derived from the 2,880 persisted E5 result rows; systems were not rerun. Every system/contract/delay cell passes its predefined tolerance.}
\label{tab:e6}
\centering
\small
\begin{tabularx}{\linewidth}{@{}lYl@{}}
\toprule
Contract & Metrics common to all systems and delay regimes & Tolerance result \\
\midrule
\localize & MRR 1.0; recall@1/3/5 1.0; ground-truth-line evidence 1.0; snippet recall 1.0 & 12/12 cells pass \\
\validatepatch & Passed accuracy 1.0; false-pass rate 0; false-fail rate 0; zero degradation versus serial & 12/12 cells pass \\
\bottomrule
\end{tabularx}
\end{table}

\section{Schedule and fault robustness}
\label{app:robustness}

\begin{table}[ht]
\caption{Held-out-input schedule and fault results. Patch configurations fail the structural prerequisite, so their semantic outcome is not applicable.}
\label{tab:robustness}
\centering
\small
\begin{tabular}{@{}llrrrl@{}}
\toprule
Contract & Run & Cases & Pass / preserve & Exposed failure & Semantic scope \\
\midrule
\localize & Schedules & 320 & 320 & 0 & B1--B3 \\
\localize & Faults & 480 & 160 & 320 & B1--B3 \\
\validatepatch & Schedules & 320 & 0 & 320 & B1 only; B2/B3 N/A \\
\validatepatch & Faults & 480 & 0 & 480 & B1 only; B2/B3 N/A \\
\bottomrule
\end{tabular}
\end{table}

Localization schedules include FIFO, criticality, concurrency one, maximum concurrency, and randomized scheduling. Accepted/preserved fault cases include normalized delayed or stale responses, subprocess nonzero exit, and corrupt provenance where the parent contract handles the condition. Exposed cases include exceptions, timeouts, empty or partial results, malformed or duplicate output, reducer exception, and cancellation. The distinction is contract-relative: a parent-visible failure is the correct outcome when silent success would violate failure refinement.

\FloatBarrier
\section{Reproducibility record}
\label{app:reproducibility}

The final protocol manifest freezes configuration, seeds, mutation registry, obligation definitions, artifact hashes, run identifiers, and environment metadata. The adjudication provenance correction records that the repository owner was the sole human adjudicator while preserving the hashes of the original frozen bundle. The main aggregate is generated from the following runs:
\begin{itemize}
  \item development validator ladder: \path{ladder-report-20260730T080836Z-901ec56c};
  \item localization certificate: \path{issue-certificate-localizeBug-20260801T193539Z-422ad544};
  \item patch certificate: \path{issue-certificate-validatePatch-20260801T193545Z-7e498a99};
  \item E5 performance: \path{run-e5-20260803T195940Z-00caf816};
  \item E6 quality derivation: \path{derive-e6-20260803T200018Z-c93df77b};
  \item B3 exploratory ablation: \path{b3-ablations-20260810T173053Z}.
\end{itemize}
E5 contains 2,880 measured rows and 24 excluded warmups, with zero failed measured cells. E6 contains 2,880 distribution rows and derives all metrics from the E5 artifact hashes. Counterexample replay reproduces 2,768/2,768 stored reason-code instances. The paper figures are generated directly from the paper-ready analysis summary and ablation JSON; vector PDFs and high-resolution PNG previews are retained with the source.

\paragraph{Evidence paths.}
\begin{itemize}
  \raggedright
  \item Paper-ready aggregate: \texttt{outputs/refinement/paper\_ready/}\allowbreak\mbox{\texttt{analysis\_summary.json}}.
  \item Primary decision record: \texttt{outputs/refinement/remediation/v1/}\allowbreak\mbox{\texttt{go\_no\_go.json}}.
  \item Final protocol manifest: \texttt{outputs/reports/202608\_verify\_agents/}\allowbreak\mbox{\texttt{final\_protocol\_manifest.json}}.
\end{itemize}
Full case-level observations, obligation results, and admission decisions remain in the validator-matrix run directories. These paths are repository-relative and are intended for the anonymized supplemental archive.

\end{document}